\documentclass[11pt,leqno]{amsart}
\usepackage{graphicx}
\usepackage{indentfirst,csquotes}
\usepackage[margin=1in]{geometry}

\usepackage{datetime}
\usepackage{wrapfig}
\usepackage{enumerate}
\usepackage{caption}
\usepackage{siunitx}
\usepackage{subcaption}
\usepackage{array,tabularx}
\usepackage{chngcntr}
\usepackage{afterpage}
\usepackage{ulem}
\usepackage{hyperref}
\usepackage{dirtytalk}
\usepackage{algorithm2e}
\usepackage{enumitem,amssymb}
\usepackage{pifont}
\usepackage{amsmath}
\usepackage{listings}
\usepackage{xcolor}
\usepackage{formal-grammar}
\usepackage{varwidth}
\usepackage{fmtcount}
\usepackage{tikz}
\usepackage[framemethod=tikz]{mdframed} 
\usepackage{cleveref}
\usepackage{microtype}
\usepackage{booktabs}

\usetikzlibrary{trees}
\usetikzlibrary{fit}
\usetikzlibrary{shapes}
\usetikzlibrary{arrows.meta, positioning, shadows}
\usetikzlibrary{chains,shapes.multipart}
\usetikzlibrary{external}
\usetikzlibrary{matrix}

\newcommand{\eg}{{e.g.}, }

\tikzstyle{rect} = [rectangle,fill=white, text centered]
\tikzstyle{database} = [draw, cylinder, shape border rotate = 90, aspect = 0.2]
\tikzstyle{thick-arrow} = [->, thick]
\tikzstyle{arrow} = [thick,->,>=stealth]
\tikzstyle{arrow-small} = [->,>=stealth]
\tikzstyle{simple-rect} = [rectangle, text centered, draw = black, inner sep=4mm]

\tikzset{
    doc/.style={draw, minimum height=4em, minimum width=3em, 
                fill=white, 
                double copy shadow={shadow xshift=4pt, 
                             shadow yshift=4pt, fill=white, draw}}
}

\tikzset{
    dcs/.style = {double copy shadow},
}

\hypersetup{
    colorlinks=true,
    filecolor=magenta,      
		citecolor=black,
    urlcolor=blue,
}

\hypersetup{linkcolor=black}

\newcommand{\src}[1]{\texttt{#1}}

\mdfdefinestyle{commandline}{
	leftmargin=10pt,
	rightmargin=10pt,
	innerleftmargin=15pt,
	middlelinecolor=black!50!white,
	middlelinewidth=2pt,
	frametitlerule=false,
	backgroundcolor=black!5!white,
	frametitle={LLM Output},
	frametitlefont={\normalfont\sffamily\color{white}\hspace{-1em}},
	frametitlebackgroundcolor=black!50!white,
	nobreak,
	singleextra={%
    }
}

\newenvironment{commandline}{
	\medskip
	\begin{mdframed}[style=commandline]
}{
	\end{mdframed}
	\medskip
}

\mdfdefinestyle{warning}{
	topline=false, bottomline=false,
	leftline=false, rightline=false,
	nobreak,
	singleextra={%
		\draw(P-|O)++(-0.5em,0)node(tmp1){};
		\draw(P-|O)++(0.5em,0)node(tmp2){};
		\fill[black,rotate around={45:(P-|O)}](tmp1)rectangle(tmp2);
		\node at(P-|O){\color{white}\scriptsize\bf !};
		\draw[very thick](P-|O)++(0,-1em)--(O);
	}
}

\newenvironment{prompt}[1][Prompt:]{ 
	\medskip
	\begin{mdframed}[style=warning]
		\noindent{\textbf{#1}}
}{
	\end{mdframed}
}

\newcommand{\sys}{\textsc{McCoy}\xspace}

\makeatletter
\def\@settitle{%
  \begin{center}%
    \normalfont\LARGE\bfseries
    \@title\par
  \end{center}%
}
\makeatother

\title[A PoC for Explainable Disease Diagnosis Using LLMs \& ASP]{A Proof-of-Concept for Explainable Disease Diagnosis\\ Using Large Language Models and\\ Answer Set Programming}

\author[Ioanna Gemou]{Ioanna Gemou}
\author[Evangelos Lamprou]{Evangelos Lamprou}
\thanks{Originally submitted as part of a university research competition on 1st June 2023. Both authors were students at the University of Patras, Greece, at the time.}

\begin{document}

\maketitle

\begin{abstract}
Accurate disease prediction is vital for timely intervention, 
effective treatment, and reducing medical complications. 
While symbolic AI has been applied in healthcare, 
its adoption remains limited due to the effort required for constructing high-quality knowledge bases.
This work introduces \sys, a framework that combines Large Language Models (LLMs) with Answer Set Programming (ASP) to overcome this barrier. 
\sys orchestrates an LLM to translate medical literature into ASP code, combines it with patient data, and processes it using an ASP solver to arrive at the final diagnosis.
This integration yields a robust, interpretable prediction framework that leverages the strengths of both paradigms. 
Preliminary results show \sys has strong performance on small-scale disease diagnosis tasks.
\end{abstract}

\section{Introduction}
Getting an accurate diagnosis quickly can make all the difference in patient outcomes, 
but building automated systems that practitioners can trust is still a major challenge~\cite{Vinarti2019}.
Symbolic AI, and in particular Answer Set Programming (ASP), provides clear logical reasoning~\cite{Brewka2011,Gelfond_Kahl_2014}
and has seen success in the medical domain~\cite{Alviano_2020}.
However, its adoption has been limited because constructing and maintaining large, 
domain-specific knowledge bases is very labor-intensive.
On the other hand, Large Language Models (LLMs) have shown strong results on tasks
such as medical question answering and clinical note summarization \cite{Singhal2022LargeLM,Singhal2023,Zheng2024LLMmed}, 
but their tendency to produce inconsistent or opaque outputs makes them risky for clinical use~\cite{aljohani2025comprehensivesurveytrustworthinesslarge,rudin2019stopexplainingblackbox,Bommasani2021}.

There is an acute need for clinical support systems that are both interpretable and widely applicable.
\sys takes a step in that direction by combining the strengths of LLMs and ASP. 
Instead of requiring experts to hand-craft large knowledge bases, 
\sys automatically converts medical literature into ASP rules. 
These rules are then enriched with patient symptoms and clinical indicators, 
and passed through an ASP solver to produce diagnostic suggestions together with logical justifications.
On public datasets, \sys reaches 95–100\% predictive accuracy on selected diseases, 
while giving transparent explanations for each diagnosis.


The paper begins with background on LLMs and ASP (\cref{sec:background}), 
then details the \sys framework (\cref{sec:methodology}), 
evaluates the approach on patient data (\cref{sec:results}), 
and concludes with a discussion of limitations and future directions (\cref{sec:conclusion}).

\section{Background}\label{sec:background}

\paragraph{\textbf{Large Language Models}}
LLMs are advanced computational systems 
designed to understand and generate human language~\cite{zhao2023survey}. 
These models can produce text that closely mirrors human linguistic patterns and have emergent reasoning capabilities~\cite{Paranjape2023}.
Recent models such as \textit{GPT-5}~\cite{gpt5}
have been trained on massive corpora spanning diverse domains, 
enabling them to acquire broad, cross-disciplinary knowledge. 
To give an intuition of the way they operate, LLMs are deep neural networks that have converged to transformer networks~\cite{Vaswani2017},
which are trained using self-supervised learning~\cite{Gui2024surveySSL}.
During training, these models refine an internal representation that enables them to analyze and understand the relationships between words, phrases, and sentences. 
By capturing patterns and structures in the training data, 
they can then be probed to generate coherent and relevant responses based on given prompts.
The term ``large" often refers to models with a large number of parameters (often in the billions)
that allow these models to build high-quality representations of language.
At a very basic level, an LLM is able to predict the next word (referred to as a token) based on its preceding context.


LLMs have demonstrated strong performance across a variety of natural language tasks, 
including question answering \cite{Brown2020}, summarization \cite{Zhang2020}, 
dialogue generation \cite{Thoppilan2022lamda}, and program synthesis \cite{Chen2021}. 
In specialized domains such as biomedicine, domain-adapted LLMs have been employed 
for clinical decision support \cite{Singhal2023} and 
biomedical knowledge retrieval \cite{Luo2022}.
Among their capabilities, LLMs excel at transforming text from one representation to another~\cite{Vaswani2017,Li2024LLMDataProcessing}.
This characteristic is particularly relevant to the present work, 
which focuses on the effective encoding of medical literature. \\

\paragraph{\textbf{Challenges in applying LLMs}}

However, LLMs are not without errors~\cite{Raj2023, Ruis2023}. 
They can occasionally produce incorrect or nonsensical responses 
\cite{Huang2025SurveyHallucinationLLM}.
There have been efforts to mitigate these issues 
through data preprocessing techniques and fine-tuning \cite{Dodge2021}.
Ongoing research seeks to improve the factual accuracy, safety, and interpretability 
of these models, enabling their deployment in high-stakes environments \cite{Ganguli2022}.
This work specifically tackles this issue by grounding LLM outputs on real medical knowledge, and having an ASP solver provide further validation and explainability.
\\

\paragraph{\textbf{Answer Set Programming}}

Answer Set Programming (ASP) \cite{Eiter2009} is a declarative problem-solving paradigm 
rooted in logic programming and non-monotonic reasoning \cite{Brewka2011}. 
The formal semantics of stable models and the foundational ASP language 
were first introduced by Gelfond and Lifschitz \cite{gel88}.
These are implemented in a family of languages commonly known as \textit{AnsProlog} \cite{Gelfond2002}.

The central idea of ASP is to represent a problem 
declaratively using logical rules, facts, and constraints. 
These elements collectively form a program that describes the conditions any solution must satisfy. 
Given a valid program, an ASP solver processes it to compute one or more stable models (also known as answer sets),
each of which corresponds to a solution that satisfies all the given rules and constraints.

\begin{figure}[t]
    \begin{center}
\begin{tikzpicture}[node distance=2.9cm]
        \node [simple-rect] (problem)  at (0,0) {Problem};
        \node [simple-rect, below of=problem] (program) {Program};
        
        \node [simple-rect, right of=program] (grounder) {Grounder};
        \node [simple-rect, right=2cm of grounder] (solver) {Solver};
        
        \node [simple-rect, right of=solver] (stable-models) {Stable Models};
        
        \node [simple-rect, above of=stable-models] (solution) {Solution};

        \draw [arrow] (problem) -- node[right] {\textbf{Modeling}} (program);
        \draw [arrow] (stable-models) -- node[left] {Interpreting} (solution);

        \draw [arrow-small] (program) -- node[above] {} (grounder);
        \draw [arrow-small] (grounder) -- node[above] {\footnotesize Grounding} (solver);
        \draw [arrow-small] (solver) -- node[above] {} (stable-models);

        \node[draw, dotted, fit=(grounder) (solver), inner sep=2mm, label=above:{Solving}] {};
\end{tikzpicture}
    \end{center}
		\caption{\textbf{Solving a problem with ASP}. A problem is first modeled as a program, then transformed into ground rules by a grounder. A solver computes the stable models, which are interpreted as solutions.}
    \label{fig:asp-solving}
\end{figure}
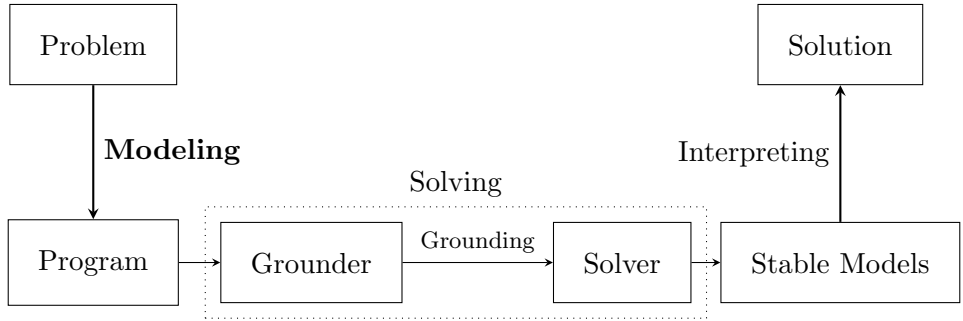

In traditional programming, transitioning from a problem to a solution 
requires the programmer's thorough understanding of the given problem, 
followed by the creation of a program that will take as input an instance of the problem,
and produce output which will then be interpreted as the solution. 
In ASP, the process of deriving solutions from a problem specification 
typically follows a sequence of well-defined steps \cite{Gebser2013}, 
as illustrated in \cref{fig:asp-solving}. 
First, the problem is modeled using the ASP language. 
Next, a grounder (\eg gringo \cite{Gebser2014}) transforms the program into a set of basic rules and facts. 
Finally, a solver (\eg clasp \cite{Holldobler2014}) computes the stable models (answer sets), 
which correspond to the solutions of the original problem.

Several solvers for ASP have been developed, with some seeing industrial adoption~\cite{Xia2020,Gebser2014}.
This work uses the Clingo ASP system,
which integrates the \src{gringo} grounder \cite{Gebser2014} and the \src{clasp} solver \cite{Holldobler2014} into a single pipeline.
The Clingo system also offers a powerful Python API, enabling integration into other applications.

\section{Related work}
\paragraph{\textbf{ASP for disease diagnosis}}
ASP has been used in the past to represent and reason over medical knowledge in a declarative, logic-based manner. 
Early work applied ASP to biomedical ontologies and structured databases to answer complex diagnostic 
and pharmacological queries with interpretable logical justifications \cite{Erdem2011}. 
In the clinical domain, ASP is especially effective for encoding medical rules and patient-specific data, 
enabling precise and explainable decision-making~\cite{Alviano_2020}

Two common strategies illustrate the role of ASP in medical reasoning. 
The first is \textit{rule-based representation}, where medical knowledge is expressed as logical statements 
that capture relationships among symptoms, diseases, treatments, and patient data. 
For instance, a rule might specify that if a patient exhibits certain symptoms, 
this supports or excludes the presence of particular diseases. 
The second is \textit{knowledge base construction}, where facts, rules, and constraints 
are combined into structured repositories that consistently model diagnostic criteria, 
symptom–disease associations, and therapeutic protocols in an extensible way. 
This work builds on these foundations and works towards closing the gap with fully automated knowledge-base construction.
\\

\paragraph{\textbf{Integrating LLMs with ASP}}

There has been work that explores the synergy between LLMs and ASP 
for enhanced diagnostic reasoning. 
These hybrid approaches aim to combine the language understanding of LLMs 
with the precision and explainability of symbolic reasoning. 
A YAML-based interface has been proposed to guide LLMs 
in extracting structured facts from natural language, 
which are then processed by ASP for logical inference \cite{alviano2024llm2asp}. 
Other work uses ASP to validate and trace misleading or inconsistent outputs 
from LLMs in medical contexts \cite{Nguyen2025}. 
Another work leverages smaller-sized LLMs and uses fine-turning to enhance their performance in generating valid ASP programs~\cite{coppolillo2024llasp}.
These works share several of the same insights and motivations as \sys.

\section{Methodology} \label{sec:methodology}

\paragraph{\textbf{Architecture}}

\begin{figure}[t]
    \includegraphics[width=\textwidth]{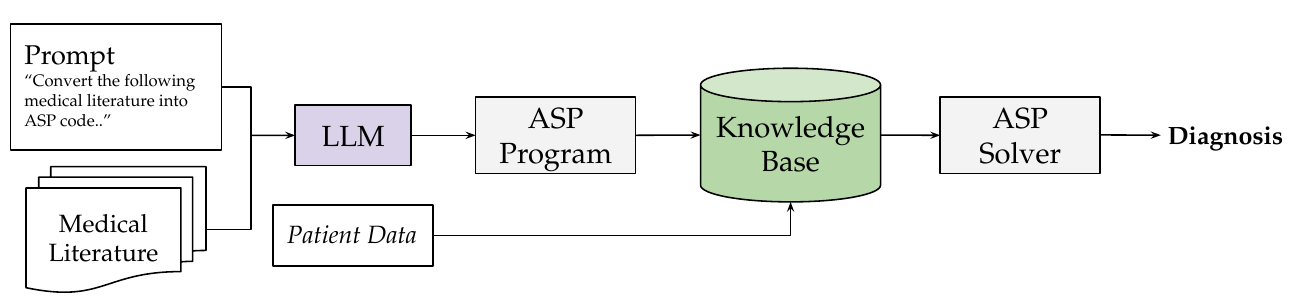}
    \caption{\textbf{\sys overview.} The diagnostic process begins by transforming medical literature 
    into an ASP program using a LLM. 
    An ASP solver then executes the program and produces a final diagnosis 
    based on each patient’s data.}
    \label{fig:architecture}
\end{figure}

\Cref{fig:architecture} illustrates \sys's architecture, 
which transforms unstructured medical text into a structured knowledge base. 
The process begins with the collection of medical literature relevant to the diseases or conditions of interest. 
A carefully designed prompt is then used to guide an LLM in translating this literature into ASP code, 
which is incorporated into a growing knowledge base. 
This knowledge base is further enriched with patient-specific information such as symptoms and examination results. 
An ASP solver subsequently reasons over the combined knowledge base and produces a set of plausible diagnoses.

\paragraph{\textbf{Prompt Construction}}
While Large Language Models exhibit strong instruction-following capabilities, 
their reliability in high-stakes domains remains limited \cite{bowman2023things}. 
To improve robustness, a wide range of prompting strategies are employed by \sys~\cite{liu2021pretrain, promptingguidePromptEngineering}.
In \sys, carefully designed prompts are employed to guide the model in translating medical literature.
This involves a careful and accurate description of the desired output format, and specific instructions on the kind of information to encode within it.
\Cref{fig:bad-prompt} represents an example of a poorly constructed prompt that fails to yield satisfactory results,
and \cref{fig:good-prompt} shows the prompt that \sys ultimately uses to convert medical text into ASP code.

\begin{figure}[!h]
\centering
\begin{prompt}
\textit{Medical Text...}

The paragraph above lists common symptoms of pneumonia.
Write a clingo script that diagnoses pneumonia based on these symptoms.
\end{prompt}

\begin{commandline}
\footnotesize
\begin{verbatim}

symptom(cough).
symptom(cough_with_mucus).
symptom(chest_pain).
symptom(headache).
...

diagnosis(pneumonia) :- symptom(cough), 
                        symptom(chest_pain), 
                        symptom(headache), 
                        symptom(arthralgia),
                        ...
\end{verbatim}
\end{commandline}

\caption{A prompt that does not yield satisfactory results. The LLM output does not decompose the disease diagnosis into discrete rules.}
\label{fig:bad-prompt}
\end{figure}

\begin{figure}[!h]
\begin{prompt}
\textit{Medical Text...}

The paragraph above lists common symptoms of pneumonia.
Write a clingo script that diagnoses pneumonia based on these symptoms.
In the diagnosis rule, use a structure like:

\src{diagnosis(pneumonia) :- has(symptom(x)), has(symptom(y)) ...}

Include alternative diagnoses that share overlapping symptoms with pneumonia.
Add rules that link one symptom to another (\eg symptom propagation or dependency).

\end{prompt}

\begin{commandline}
\footnotesize
\begin{verbatim}
symptom(cough).
symptom(cough_with_mucus).
...


linked_symptom(cough_with_mucus, wheezing).
linked_symptom(cough_with_mucus, shortness_of_breath).

...

has(symptom(Symptom2)) :- has(symptom(Symptom1)), 
              linked_symptom(Symptom1, Symptom2).


diagnosis(pneumonia) :- has(symptom(cough)),
                        has(symptom(cough_with_mucus)),
...

diagnosis(pneumonia) :- has(symptom(cough_with_mucus)),
                        has(symptom(chest_pain)),
...
\end{verbatim}
\end{commandline}
\caption{An effective prompt that produces accurate and well-structured output.
	The LLM output is structured into multiple rules that aid explainability and can result in partial diagnoses.}
\label{fig:good-prompt}
\end{figure}

\paragraph{\textbf{Logical Program Construction}}

The knowledge base is initialized with a set of symptoms and medical indicators represented as:
\begin{align}
    symptom(s). & \; s \in \{ \text{cough, chest pain, rash}, \dots \} \\
    indicator(i). & \; i \in \{ \text{low MCV, high TIBC}, \dots \}
\end{align}
Let $S$ denote the set of symptoms and $I$ the set of indicators.
A patient exhibiting a specific symptom or testing positive for an indicator is represented as:
\begin{equation}
    has(x). \; x \in S \cup I
\end{equation}

Logical rules used for inferring diagnoses follow this general form:
\begin{align}
    diagnosis(d_1) & \longleftarrow has(x_1) \land has(x_2) \land \dots \\
    diagnosis(d_1) & \longleftarrow has(x_1) \land has(x_3) \land \dots \\
    diagnosis(d_2) & \longleftarrow has(x_3) \land has(x_4) \land \dots
\end{align}

Correlations between certain symptoms are also encoded in the knowledge base. 
For example, clinical observations have reported links between skin conditions and headaches \cite{migraine-hives}. 
In some cases, the presence of one symptom may lead to the appearance of another. 
Such relationships are captured using rules of the form:
\begin{align}
    has(symptom(S2)) \longleftarrow has(symptom(S1)) \land linked\_symptom(S1, S2).
\end{align}

Example entries for such relationships include:
\begin{align}
    linked\_symptom(grunting, chest\_retractions).\\
    linked\_symptom(nasal\_flaring, chest\_retractions).
\end{align}

Since patients may not exhibit \textit{all} known symptoms, 
\sys must support reasoning under partial information. 
To allow flexible reasoning, a choice rule introduces possible symptoms 
into the knowledge base when patient data is incomplete:
\begin{equation}
    \{ add(symptom(S)) : symptom(S) \}.
\end{equation}

The following integrity constraint imposes at least one diagnosis:
\begin{equation}
    \bot \longleftarrow \sim diagnosis(\_)
\end{equation}

Finally, the solver minimizes assumptions by penalizing added symptoms, 
favoring diagnoses based on known patient information:
\begin{equation}
    \#minimize \{ 1, S : add(symptom(S)) \}
\end{equation}

This formulation guarantees that at least one diagnosis is produced, 
relying as much as possible on available clinical data rather than hypothetical assumptions.
\\

\paragraph{\textbf{Diagnostic Explainability}}

\begin{figure}[t]
    \centering
    \begin{minipage}{0.48\textwidth}
    \begin{equation}
    \begin{aligned}[b]
        l :& punish \longleftarrow drive, drunk \\
        m :& punish \longleftarrow resist \\
        e :& prison \longleftarrow punish \\
        d :& drive \\
        k :& drunk \\
        r :& resist \\
    \end{aligned}
    \label{eq:example-cg}
    \end{equation}
    \end{minipage}
    \hfill
    \begin{minipage}{0.45\textwidth}
    \centering
    \includegraphics[width=.5\textwidth]{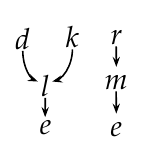}
    \caption{Causal graph of logic program in \cref{eq:example-cg}.}
    \label{fig:causal-g}
    \end{minipage}
\end{figure}

Symbolic AI offers a significant advantage 
over sub-symbolic methods in terms of explainability. 
In symbolic systems, the relationships between facts and conclusions are explicitly defined, 
enabling transparent reasoning.
Thus, it possible to associate logical propositions with their justifications~\cite{cabalar2014causal}.
For example, 
the logic program shown in \cref{eq:example-cg} is accompanied by a causal graph 
shown in \cref{fig:causal-g}, which illustrates the connections 
between rules and their inferred outcomes. 
This structure supports the interpretability of conclusions derived by a symbolic logic solver.
The tool \src{xclingo}~\cite{Cabalar_2020} 
enables the visualization of justifications for results produced by the \src{clingo} solver. 
It does so by incorporating a trace that corresponds to the rules of the logic program. 
\sys leverages this produce such a trace, which can then help identify the reasoning path, as well as potential flaws in \sys's knowledge base and reasoning process.

\section{Results}\label{sec:results}

Evaluation was conducted using a publicly available dataset.\footnote{\url{https://www.kaggle.com/datasets/itachi9604/disease-symptom-description-dataset}}
Each record's list of symptoms serves as input.  
The knowledge base—comprising ASP rules automatically derived  
from medical literature—infers potential diagnoses.  
A prediction is considered correct when the disease inferred  
by the ASP solver matches the actual diagnosis listed in the dataset.

\begin{table}[h]
    \centering
    \caption{Accuracy of the framework on selected diseases. 
    The size column indicates the number of logical terms in the disease-specific ASP program.}
    \begin{tabular}{lcc}
    \toprule
    Diseases & Size & Accuracy \\
    \midrule
    Chickenpox     &  66  & 95\% \\
    Pneumonia      &  75  & 100\% \\
    Common Cold    &  44  & 100\% \\
    \bottomrule
    \end{tabular}
\end{table}

High accuracy is achieved when the knowledge base includes a sufficiently 
rich set of logical rules.
In addition to accurate predictions, \sys offers explainability and visualizes the reasoning process behind each diagnosis~(an example is shown in \cref{lst:chickenpox-explanation}).

\begin{lstlisting}[caption={Excerpt from the explanation tree for the chickenpox diagnosis. The structure illustrates the reasoning path based on symptom associations (slightly modified for clarity).},
    label=lst:chickenpox-explanation]{Name}
    *
    |__ diagnosis(chickenpox)
        |__ has(symptom(itching))
        |__ has(symptom(fatigue))
        |__ has(symptom(lethargy))
            |__ has(symptom(fatigue))
            |__ linked_symptom(fatigue, lethargy)
        |__ has(symptom(high_fever))
            |__ has(symptom(mild_fever))
                |__ has(symptom(loss_of_appetite))
                |__ linked_symptom(loss_of_appetite, mild_fever)
            |__ linked_symptom(mild_fever, high_fever)
        |__ has(symptom(loss_of_appetite))
        |__ has(symptom(mild_fever))
            |__ has(symptom(loss_of_appetite))
            |__ linked_symptom(loss_of_appetite, mild_fever)
        |__ has(symptom(swelled_lymph_nodes))
\end{lstlisting}
    
\Cref{lst:chickenpox-explanation} shows an excerpt from the explanation 
tree generated for a patient diagnosed with \textit{chickenpox}. 
The tree visualizes the reasoning steps performed by the ASP solver to reach this conclusion. 
At the root, \sys infers \texttt{diagnosis(chickenpox)} 
based on the presence of multiple symptoms such as \texttt{itching}, \texttt{fatigue}, \texttt{lethargy}, 
and \texttt{high\_fever}. 
Each symptom is further justified through observed inputs or inferred relationships. 
For example, \texttt{lethargy} is supported by repeated instances of \texttt{fatigue} 
and the known association \texttt{linked\_symptom(fatigue, lethargy)}. 
Similarly, \texttt{high\_fever} is explained via the intermediate symptom \texttt{mild\_fever}, 
which itself is supported by \texttt{loss\_of\_appetite}. 
These explanations demonstrate how \sys combines direct observations 
and transitive symptom relationships encoded in the knowledge base to 
construct a transparent and explainable diagnostic path.

\section{Conclusion}\label{sec:conclusion}

This paper presents \sys, a framework for disease prediction using LLMs and ASP. 
The system uses LLMs to extract knowledge from medical texts and translate it into ASP rules, 
which then drive predictions and provide interpretable explanations. 
The architecture, prompt design, and rule structuring in the knowledge base are described.
Experiments show that \sys can predict diseases accurately 
while keeping the reasoning process transparent.

Future work will focus on scaling this process across more sources, 
formalizing the pipeline to enable auditability and pluggability,
and further experimentation in augmenting each step with additional guardrails, possibly including fine-tuning.
\newpage

\bibliographystyle{unsrt}
\bibliography{bib}

\begin{thebibliography}{10}

\bibitem{Vinarti2019}
Retno~Aulia Vinarti.
\newblock {Knowledge Representation for Infectious Disease Risk Prediction
  System: A Literature Review}.
\newblock {\em {Procedia Computer Science}}, 161:821--825, 2019.

\bibitem{Brewka2011}
Gerhard Brewka, Thomas Eiter, and Miros\l{}aw Truszczy\'{n}ski.
\newblock {Answer set programming at a glance}.
\newblock {\em {Commun. ACM}}, 54(12):92–103, December 2011.

\bibitem{Gelfond_Kahl_2014}
Michael Gelfond and Yulia Kahl.
\newblock {\em Knowledge Representation, Reasoning, and the Design of
  Intelligent Agents: The Answer-Set Programming Approach}.
\newblock Cambridge University Press, 2014.

\bibitem{Alviano_2020}
Mario Alviano, Riccardo Bertolucci, Valeria Cardellini, Carmine Dodaro,
  Giuseppe Galatà, Muhammad~Kashif Khan, Nicola Leone, Marco Maratea,
  Francesco Ricca, and Marco Schouten.
\newblock {Answer Set Programming in Healthcare: Extended Overview}.
\newblock {\em {IPS-RCRA@ AI* IA}}, 2020.

\bibitem{Singhal2022LargeLM}
K.~Singhal, Shekoofeh Azizi, Tao Tu, Said Mahdavi, Jason Wei, Hyung~Won Chung,
  Nathan Scales, Ajay~Kumar Tanwani, Heather~J. Cole-Lewis, Stephen~J. Pfohl,
  P~A Payne, Martin~G. Seneviratne, Paul Gamble, Chris Kelly, Nathaneal
  Scharli, Aakanksha Chowdhery, P.~A. Mansfield, Blaise~Ag{\"u}era y~Arcas,
  Dale~R. Webster, Greg~S. Corrado, Yossi Matias, Katherine Hui-Ling Chou,
  Juraj Gottweis, Nenad Tomaev, Yun Liu, Alvin Rajkomar, Jo{\"e}lle~K. Barral,
  Christopher Semturs, Alan Karthikesalingam, and Vivek Natarajan.
\newblock {Large language models encode clinical knowledge}.
\newblock {\em {Nature}}, 620:172--180, 2022.

\bibitem{Singhal2023}
Karan Singhal, Tao Tu, Juraj Gottweis, Rory Sayres, Ellery Wulczyn, Le~Hou,
  Kevin Clark, Stephen Pfohl, Heather Cole-Lewis, Darlene Neal, Mike
  Schaekermann, Amy Wang, Mohamed Amin, Sami Lachgar, Philip Mansfield, Sushant
  Prakash, Bradley Green, Ewa Dominowska, Blaise~Aguera y~Arcas, Nenad Tomasev,
  Yun Liu, Renee Wong, Christopher Semturs, S.~Sara Mahdavi, Joelle Barral,
  Dale Webster, Greg~S. Corrado, Yossi Matias, Shekoofeh Azizi, Alan
  Karthikesalingam, and Vivek Natarajan.
\newblock {Towards Expert-Level Medical Question Answering with Large Language
  Models}, 2023.

\bibitem{Zheng2024LLMmed}
Yanxin Zheng, Wensheng Gan, Zefeng Chen, Zhenlian Qi, Qian Liang, and Philip~S.
  Yu.
\newblock {Large Language Models for Medicine: A Survey}, 2024.

\bibitem{aljohani2025comprehensivesurveytrustworthinesslarge}
Manar Aljohani, Jun Hou, Sindhura Kommu, and Xuan Wang.
\newblock {A Comprehensive Survey on the Trustworthiness of Large Language
  Models in Healthcare}, 2025.

\bibitem{rudin2019stopexplainingblackbox}
Cynthia Rudin.
\newblock {Stop Explaining Black Box Machine Learning Models for High Stakes
  Decisions and Use Interpretable Models Instead}, 2019.

\bibitem{Bommasani2021}
Rishi Bommasani, Drew~A. Hudson, Ehsan Adeli, Russ Altman, Simran Arora, Sydney
  von Arx, Michael~S. Bernstein, Jeannette Bohg, Antoine Bosselut, Emma
  Brunskill, Erik Brynjolfsson, Shyamal Buch, Dallas Card, Rodrigo Castellon,
  Niladri Chatterji, Annie Chen, Kathleen Creel, Jared~Quincy Davis, Dora
  Demszky, Chris Donahue, Moussa Doumbouya, Esin Durmus, Stefano Ermon, John
  Etchemendy, Kawin Ethayarajh, Li~Fei-Fei, Chelsea Finn, Trevor Gale, Lauren
  Gillespie, Karan Goel, Noah Goodman, Shelby Grossman, Neel Guha, Tatsunori
  Hashimoto, Peter Henderson, John Hewitt, Daniel~E. Ho, Jenny Hong, Kyle Hsu,
  Jing Huang, Thomas Icard, Saahil Jain, Dan Jurafsky, Pratyusha Kalluri,
  Siddharth Karamcheti, Geoff Keeling, Fereshte Khani, Omar Khattab, Pang~Wei
  Koh, Mark Krass, Ranjay Krishna, Rohith Kuditipudi, Ananya Kumar, Faisal
  Ladhak, Mina Lee, Tony Lee, Jure Leskovec, Isabelle Levent, Xiang~Lisa Li,
  Xuechen Li, Tengyu Ma, Ali Malik, Christopher~D. Manning, Suvir Mirchandani,
  Eric Mitchell, Zanele Munyikwa, Suraj Nair, Avanika Narayan, Deepak
  Narayanan, Ben Newman, Allen Nie, Juan~Carlos Niebles, Hamed Nilforoshan,
  Julian Nyarko, Giray Ogut, Laurel Orr, Isabel Papadimitriou, Joon~Sung Park,
  Chris Piech, Eva Portelance, Christopher Potts, Aditi Raghunathan, Rob Reich,
  Hongyu Ren, Frieda Rong, Yusuf Roohani, Camilo Ruiz, Jack Ryan, Christopher
  Ré, Dorsa Sadigh, Shiori Sagawa, Keshav Santhanam, Andy Shih, Krishnan
  Srinivasan, Alex Tamkin, Rohan Taori, Armin~W. Thomas, Florian Tramèr,
  Rose~E. Wang, William Wang, Bohan Wu, Jiajun Wu, Yuhuai Wu, Sang~Michael Xie,
  Michihiro Yasunaga, Jiaxuan You, Matei Zaharia, Michael Zhang, Tianyi Zhang,
  Xikun Zhang, Yuhui Zhang, Lucia Zheng, Kaitlyn Zhou, and Percy Liang.
\newblock {On the Opportunities and Risks of Foundation Models}, 2021.

\bibitem{zhao2023survey}
Wayne~Xin Zhao, Kun Zhou, Junyi Li, Tianyi Tang, Xiaolei Wang, Yupeng Hou,
  Yingqian Min, Beichen Zhang, Junjie Zhang, Zican Dong, Yifan Du, Chen Yang,
  Yushuo Chen, Zhipeng Chen, Jinhao Jiang, Ruiyang Ren, Yifan Li, Xinyu Tang,
  Zikang Liu, Peiyu Liu, Jian-Yun Nie, and Ji-Rong Wen.
\newblock {A Survey of Large Language Models}, 2023.

\bibitem{Paranjape2023}
Bhargavi Paranjape, Scott Lundberg, Sameer Singh, Hannaneh Hajishirzi, Luke
  Zettlemoyer, and Marco~Tulio Ribeiro.
\newblock {ART: Automatic multi-step reasoning and tool-use for large language
  models}, 2023.

\bibitem{gpt5}
OpenAI.
\newblock {GPT-5}, 2025.

\bibitem{Vaswani2017}
Ashish Vaswani, Noam Shazeer, Niki Parmar, Jakob Uszkoreit, Llion Jones,
  Aidan~N Gomez, \L~ukasz Kaiser, and Illia Polosukhin.
\newblock {Attention is All you Need}.
\newblock In I.~Guyon, U.~Von Luxburg, S.~Bengio, H.~Wallach, R.~Fergus,
  S.~Vishwanathan, and R.~Garnett, editors, {\em Advances in Neural Information
  Processing Systems}, volume~30. Curran Associates, Inc., 2017.

\bibitem{Gui2024surveySSL}
Jie Gui, Tuo Chen, Jing Zhang, Qiong Cao, Zhenan Sun, Hao Luo, and Dacheng Tao.
\newblock A survey on self-supervised learning: Algorithms, applications, and
  future trends.
\newblock {\em IEEE Trans. Pattern Anal. Mach. Intell.}, 46(12):9052–9071,
  December 2024.

\bibitem{Brown2020}
Tom Brown, Benjamin Mann, Nick Ryder, Melanie Subbiah, Jared~D Kaplan, Prafulla
  Dhariwal, Arvind Neelakantan, Pranav Shyam, Girish Sastry, Amanda Askell,
  Sandhini Agarwal, Ariel Herbert-Voss, Gretchen Krueger, Tom Henighan, Rewon
  Child, Aditya Ramesh, Daniel Ziegler, Jeffrey Wu, Clemens Winter, Chris
  Hesse, Mark Chen, Eric Sigler, Mateusz Litwin, Scott Gray, Benjamin Chess,
  Jack Clark, Christopher Berner, Sam McCandlish, Alec Radford, Ilya Sutskever,
  and Dario Amodei.
\newblock {Language Models are Few-Shot Learners}.
\newblock In H.~Larochelle, M.~Ranzato, R.~Hadsell, M.F. Balcan, and H.~Lin,
  editors, {\em Advances in Neural Information Processing Systems}, volume~33,
  pages 1877--1901. Curran Associates, Inc., 2020.

\bibitem{Zhang2020}
Jingqing Zhang, Yao Zhao, Mohammad Saleh, and Peter Liu.
\newblock {{PEGASUS}: Pre-training with Extracted Gap-sentences for Abstractive
  Summarization}.
\newblock In Hal~Daumé III and Aarti Singh, editors, {\em Proceedings of the
  37th International Conference on Machine Learning}, volume 119 of {\em
  Proceedings of Machine Learning Research}, pages 11328--11339. PMLR, 13--18
  Jul 2020.

\bibitem{Thoppilan2022lamda}
Romal Thoppilan, Daniel~De Freitas, Jamie Hall, Noam Shazeer, Apoorv
  Kulshreshtha, Heng-Tze Cheng, Alicia Jin, Taylor Bos, Leslie Baker, Yu~Du,
  YaGuang Li, Hongrae Lee, Huaixiu~Steven Zheng, Amin Ghafouri, Marcelo
  Menegali, Yanping Huang, Maxim Krikun, Dmitry Lepikhin, James Qin, Dehao
  Chen, Yuanzhong Xu, Zhifeng Chen, Adam Roberts, Maarten Bosma, Vincent Zhao,
  Yanqi Zhou, Chung-Ching Chang, Igor Krivokon, Will Rusch, Marc Pickett,
  Pranesh Srinivasan, Laichee Man, Kathleen Meier-Hellstern, Meredith~Ringel
  Morris, Tulsee Doshi, Renelito~Delos Santos, Toju Duke, Johnny Soraker, Ben
  Zevenbergen, Vinodkumar Prabhakaran, Mark Diaz, Ben Hutchinson, Kristen
  Olson, Alejandra Molina, Erin Hoffman-John, Josh Lee, Lora Aroyo, Ravi
  Rajakumar, Alena Butryna, Matthew Lamm, Viktoriya Kuzmina, Joe Fenton, Aaron
  Cohen, Rachel Bernstein, Ray Kurzweil, Blaise Aguera-Arcas, Claire Cui,
  Marian Croak, Ed~Chi, and Quoc Le.
\newblock {LaMDA: Language Models for Dialog Applications}, 2022.

\bibitem{Chen2021}
Mark Chen, Jerry Tworek, Heewoo Jun, Qiming Yuan, Henrique~Ponde
  de~Oliveira~Pinto, Jared Kaplan, Harri Edwards, Yuri Burda, Nicholas Joseph,
  Greg Brockman, Alex Ray, Raul Puri, Gretchen Krueger, Michael Petrov, Heidy
  Khlaaf, Girish Sastry, Pamela Mishkin, Brooke Chan, Scott Gray, Nick Ryder,
  Mikhail Pavlov, Alethea Power, Lukasz Kaiser, Mohammad Bavarian, Clemens
  Winter, Philippe Tillet, Felipe~Petroski Such, Dave Cummings, Matthias
  Plappert, Fotios Chantzis, Elizabeth Barnes, Ariel Herbert-Voss,
  William~Hebgen Guss, Alex Nichol, Alex Paino, Nikolas Tezak, Jie Tang, Igor
  Babuschkin, Suchir Balaji, Shantanu Jain, William Saunders, Christopher
  Hesse, Andrew~N. Carr, Jan Leike, Josh Achiam, Vedant Misra, Evan Morikawa,
  Alec Radford, Matthew Knight, Miles Brundage, Mira Murati, Katie Mayer, Peter
  Welinder, Bob McGrew, Dario Amodei, Sam McCandlish, Ilya Sutskever, and
  Wojciech Zaremba.
\newblock {Evaluating Large Language Models Trained on Code}, 2021.

\bibitem{Luo2022}
Renqian Luo, Liai Sun, Yingce Xia, Tao Qin, Sheng Zhang, Hoifung Poon, and
  Tie-Yan Liu.
\newblock {BioGPT: generative pre-trained transformer for biomedical text
  generation and mining}.
\newblock {\em {Briefings in Bioinformatics}}, 23(6), September 2022.

\bibitem{Li2024LLMDataProcessing}
Bohang Li, Gaozhe Jiang, Ningxin Li, and Chaoda Song.
\newblock Research on large-scale structured and unstructured data processing
  based on large language model.
\newblock In {\em Proceedings of the International Conference on Machine
  Learning, Pattern Recognition and Automation Engineering}, MLPRAE '24, page
  111–116, New York, NY, USA, 2024. Association for Computing Machinery.

\bibitem{Raj2023}
Harsh Raj, Domenic Rosati, and Subhabrata Majumdar.
\newblock {Measuring Reliability of Large Language Models through Semantic
  Consistency}, 2023.

\bibitem{Ruis2023}
Laura~Eline Ruis, Akbir Khan, Stella Biderman, Sara Hooker, Tim
  Rockt{\"a}schel, and Edward Grefenstette.
\newblock {Large language models are not zero-shot communicators}, 2023.

\bibitem{Huang2025SurveyHallucinationLLM}
Lei Huang, Weijiang Yu, Weitao Ma, Weihong Zhong, Zhangyin Feng, Haotian Wang,
  Qianglong Chen, Weihua Peng, Xiaocheng Feng, Bing Qin, and Ting Liu.
\newblock A survey on hallucination in large language models: Principles,
  taxonomy, challenges, and open questions.
\newblock {\em ACM Trans. Inf. Syst.}, 43(2), January 2025.

\bibitem{Dodge2021}
Jesse Dodge, Maarten Sap, Ana Marasović, William Agnew, Gabriel Ilharco, Dirk
  Groeneveld, Margaret Mitchell, and Matt Gardner.
\newblock {Documenting Large Webtext Corpora: A Case Study on the Colossal
  Clean Crawled Corpus}, 2021.

\bibitem{Ganguli2022}
Deep Ganguli, Liane Lovitt, Jackson Kernion, Amanda Askell, Yuntao Bai, Saurav
  Kadavath, Ben Mann, Ethan Perez, Nicholas Schiefer, Kamal Ndousse, Andy
  Jones, Sam Bowman, Anna Chen, Tom Conerly, Nova DasSarma, Dawn Drain, Nelson
  Elhage, Sheer El-Showk, Stanislav Fort, Zac Hatfield-Dodds, Tom Henighan,
  Danny Hernandez, Tristan Hume, Josh Jacobson, Scott Johnston, Shauna Kravec,
  Catherine Olsson, Sam Ringer, Eli Tran-Johnson, Dario Amodei, Tom Brown,
  Nicholas Joseph, Sam McCandlish, Chris Olah, Jared Kaplan, and Jack Clark.
\newblock {Red Teaming Language Models to Reduce Harms: Methods, Scaling
  Behaviors, and Lessons Learned}, 2022.

\bibitem{Eiter2009}
Thomas Eiter, Giovambattista Ianni, and Thomas Krennwallner.
\newblock {Answer Set Programming: A Primer}.
\newblock volume 5689, pages 40--110, 01 2009.

\bibitem{gel88}
Michael Gelfond and Vladimir Lifschitz.
\newblock {The Stable Model Semantics for Logic Programming}.
\newblock In Robert Kowalski, Bowen, and Kenneth, editors, {\em Proceedings of
  International Logic Programming Conference and Symposium}, pages 1070--1080.
  MIT Press, 1988.

\bibitem{Gelfond2002}
Michael Gelfond.
\newblock {Representing Knowledge in A-Prolog}.
\newblock In {\em Computational Logic: Logic Programming and Beyond}, pages
  413--451. Springer Berlin Heidelberg, 2002.

\bibitem{Gebser2013}
Martin Gebser, Roland Kaminski, Benjamin Kaufmann, and Torsten Schaub.
\newblock {\em {Answer Set Solving in Practice}}.
\newblock Springer International Publishing, 2013.

\bibitem{Gebser2014}
Martin Gebser, Roland Kaminski, Benjamin Kaufmann, and Torsten Schaub.
\newblock {Clingo = ASP + Control: Preliminary Report}, 2014.

\bibitem{Holldobler2014}
Steffen Hölldobler and Lukas Schweizer.
\newblock {Answer set programming and CLASP: A tutorial}.
\newblock volume 1145, 04 2014.

\bibitem{Xia2020}
Boming Xia, Xiaozhen Ye, and Adnan Abuassba.
\newblock {Recent Research on AI in Games}.
\newblock pages 505--510, 06 2020.

\bibitem{Erdem2011}
Esra Erdem, Yelda Erdem, Halit Erdogan, and Umut Oztok.
\newblock {Finding Answers and Generating Explanations for Complex Biomedical
  Queries}.
\newblock {\em {Proceedings of the {AAAI} Conference on Artificial
  Intelligence}}, 25(1):785--790, August 2011.

\bibitem{alviano2024llm2asp}
Mario Alviano and Luca Grillo.
\newblock {Answer Set Programming and Large Language Models Interaction with
  YAML: Preliminary Report}.
\newblock In {\em Proceedings of the 39th Italian Conference on Computational
  Logic (CILC 2024)}, volume 3733 of {\em CEUR Workshop Proceedings}.
  CEUR-WS.org, 2024.

\bibitem{Nguyen2025}
Quang-Anh Nguyen, Thu-Trang Pham, Thi-Hai-yen Vuong, Giang Trinh, and Nguyen
  Thanh.
\newblock {Detecting Misleading Information with LLMs and Explainable ASP}.
\newblock pages 1327--1334, 01 2025.

\bibitem{coppolillo2024llasp}
Erica Coppolillo, Francesco Calimeri, Giuseppe Manco, Simona Perri, and
  Francesco Ricca.
\newblock {LLASP: Fine-tuning Large Language Models for Answer Set
  Programming}, 2024.

\bibitem{bowman2023things}
Samuel~R. Bowman.
\newblock {Eight Things to Know about Large Language Models}, 2023.

\bibitem{liu2021pretrain}
Pengfei Liu, Weizhe Yuan, Jinlan Fu, Zhengbao Jiang, Hiroaki Hayashi, and
  Graham Neubig.
\newblock {Pre-train, Prompt, and Predict: A Systematic Survey of Prompting
  Methods in Natural Language Processing}, 2021.

\bibitem{promptingguidePromptEngineering}
{{P}rompt {E}ngineering {G}uide}.
\newblock \url{https://www.promptingguide.ai}.
\newblock [Accessed 27-May-2023].

\bibitem{migraine-hives}
{Migraine.com Editorial Team}.
\newblock {Migraine Hives Symptoms}.
\newblock \url{https://migraine.com/migraine-symptoms/hives}.
\newblock Accessed on May 30, 2023.

\bibitem{cabalar2014causal}
Pedro Cabalar, Jorge Fandinno, and Michael Fink.
\newblock {Causal Graph Justifications of Logic Programs}, 2014.

\bibitem{Cabalar_2020}
Pedro Cabalar, Jorge Fandinno, and Brais Mu{\~{n}}iz.
\newblock {A System for Explainable Answer Set Programming}.
\newblock {\em {Electronic Proceedings in Theoretical Computer Science}},
  325:124--136, sep 2020.

\end{thebibliography}

\end{document}